\documentclass{article}

\usepackage{PRIMEarxiv}

\usepackage{times}
\usepackage{latexsym}
\usepackage{inconsolata}

\usepackage{multirow}
\usepackage{tabularx}
\usepackage{natbib}
\usepackage{easyReview}

\usepackage[utf8]{inputenc} 
\usepackage[T1]{fontenc}    
\usepackage{hyperref}       
\usepackage{url}            
\usepackage{booktabs}       
\usepackage{amsfonts}       
\usepackage{nicefrac}       
\usepackage{microtype}      
\usepackage{lipsum}
\usepackage{fancyhdr}       
\usepackage{graphicx}       
\graphicspath{{media/}}     

\title{The Nuts and Bolts of Natural Language to SQL Translation: A Systematic Analysis of Model Pipeline Optimisation Approaches and their Interactions}

\author{Filip Klubi\v{c}ka \\
ADAPT Research Centre \\
Trinity College Dublin \\
Dublin, Ireland \\
\texttt{fklubicka@gmail.com} \\
  \And
Vasudevan Nedumpozhimana \\
ADAPT Research Centre \\
Trinity College Dublin \\
Dublin, Ireland \\
\texttt{vnedumpo@tcd.ie} \\
  \And
Sneha Rautmare \\
ADAPT Research Centre \\
Trinity College Dublin \\
Dublin, Ireland \\
\texttt{sneha.rautmare@adaptcentre.ie} \\
  \And
Bora Caglayan \\
Huawei Ireland Research Centre \\
Dublin, Ireland \\
\texttt{bora.caglayan@huawei.com} \\
  \And
Mingxue Wang \\
Huawei Ireland Research Centre \\
Dublin, Ireland \\
\texttt{wangmingxue1@huawei.com} \\
  \And
John D. Kelleher \\
ADAPT Research Centre \\
Trinity College Dublin \\
Dublin, Ireland \\
\texttt{john.kelleher@tcd.ie} \\
}

\begin{document}
\maketitle

\begin{abstract}
In the age of large language models, Natural Language to SQL (NL2SQL) translation remains an open problem with many useful applications. We explore interactions between several NL2SQL pipeline extensions to inspire development of more lightweight models. Specifically, we integrate the NatSQL intermediate representation, include a preprocessing step and a fine-tuning step based on synthetic data, and develop a novel reranker model to improve SQL selection in the final beam. We perform an ablation study supplemented by a Shapley analysis of these different components integrated with two backbone architectures, SmBoP and RASAT. We find that simply combining all of them does not lead to best results, but that their impact depends on their interactions with the baseline system, as well as each other.

\end{abstract}


\keywords{NL2SQL \and SQL \and reranker \and Spider \and generative models \and data augmentation \and optimization methods \and model architectures \and code generation}

\section{Introduction}

Natural Language to SQL translation (NL2SQL) is a subtask of the broader problem of semantic parsing, defined as understanding the meaning of natural language utterances and mapping them to meaningful executable queries. Solving NL2SQL is a problem with practical applications such as allowing integration of natural language interfaces to relational database management systems, enhancing accessibility and improving user experience. In an NL2SQL scenario, given a relational database and a natural language question (NLQ), the goal is to find an equivalent SQL query which will answer the NLQ once executed. The core challenge stems from understanding the meaning and intention of the NLQ, which can be expressed in myriad ways due to the high complexity and expressiveness of natural language, and translating that meaning into an SQL query, which is comparably more narrow in scope and follows significantly simpler and more rigidly defined syntactic and semantic rules. Thus NL2SQL is still considered an open problem. 


\begin{figure}[h!]
	\centering
        \includegraphics[width=1\columnwidth]{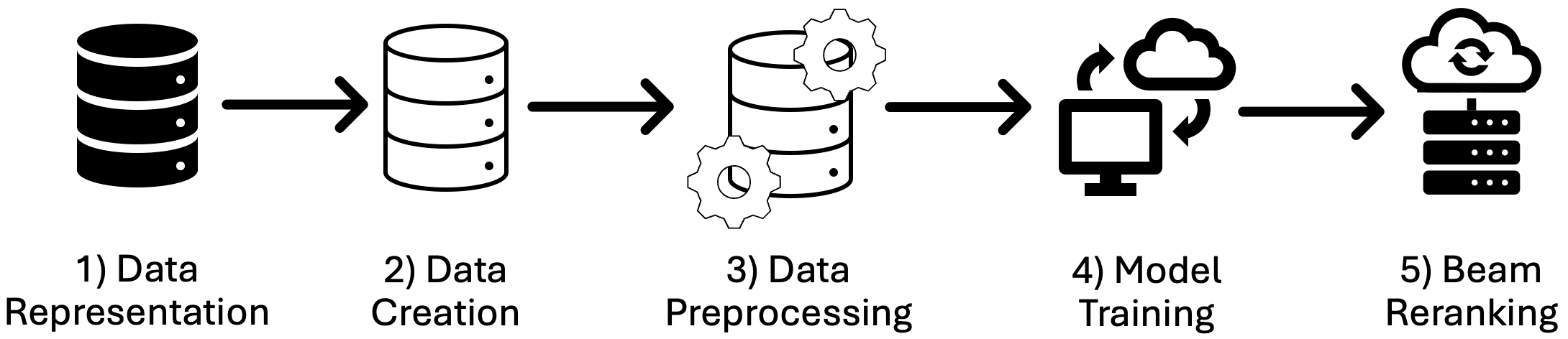}
	\caption{Points of optimisation in an NL2SQL pipeline.
    \label{f:pipeline}}
\end{figure}

As LLMs grow ever larger and more capable, it becomes more tempting to apply them to any given task. However this is not always necessary, feasible or responsible in many real-world applications or production environments, with the models being either too large, too expensive or too power-hungry to realistically deploy. We argue that domain-aligned, specialised NL2SQL systems have yet to plateau, and we explore ways to improve performance of more light-weight models. To this end, we carry out a systematic analysis of several NL2SQL pipeline extensions which have been shown to improve performance so as to understand their relative benefit as well as the benefits of combining them. Each of these extensions was selected to represent a state of the art intervention at a different stage in the NL2SQL pipeline, as shown in Figure \ref{f:pipeline}. We study the following extensions: using intermediate representations, synthetic data creation, data preprocessing, and applying a final beam reranker to select the most likely SQL candidate. These have been studied individually, but never in combination---in doing so, we discover some interesting interactions, with both positive and negative effects on overall model performance, and identify combinations that yield the largest performance improvements.

This paper's main contributions are: (a) a systematic study of interactions between SotA NL2SQL extensions; (b) the development of a novel NL2SQL final beam reranker; (c) the integration of NatSQL with RASAT and SmBoP; and (d) a Shapley analysis quantifying the contribution of individual components in combination. We find that the strongest contributors are NatSQL and our own final beam reranker, and show that the right combination of methods can dramatically raise the performance ceiling of baseline systems. 

\section{Related Work}


There is a long history for Natural Language to SQL translation research \citep{kim-2021-nl2sql} starting from rule-based methods \cite{zelle1996learning}. Earlier rule-based methods uses templates to generate SQL queries in narrowly-defined use cases, with the drawbacks of scalability and generalisability \cite{saha2016athena}. Later sequence-to-sequence architecture based approaches explored various methods such as: sketch-based decoders \citep{dong-lapata-2016-language,zhong2017seq2sql,xu2017sqlnet}, pointer-generator decoders \citep{see-etal-2017-get,hwang2019comprehensive,lin-etal-2020-bridging,hui2021improving}, graph neural networks \citep{wang-etal-2020-rat,cao-etal-2021-lgesql}, and fine-tuning pre-trained language models \citep{raffel2020exploring,scholak-etal-2021-picard,zhang2023resdsql}, and most recently Large Language Models \cite{gao2023texttosql,pourreza2023dinsql,dong2023c3,sun2023sqlpalm,liu2023comprehensive,wang2025macsqlmultiagentcollaborativeframework,heidari2025agentiqlagentinspiredmultiexpertframework}.

While recent LLM-based systems increasingly rely on modular and agentic pipelines that decompose the task into multiple stages \cite{heidari2025agentiqlagentinspiredmultiexpertframework}, our work demonstrates the effectiveness of a similar decomposition-oriented paradigm in a non-LLM setting.
Specifically, we formulate NL2SQL as a pipeline comprising preprocessing, synthetic data generation, candidate generation, and reranking stages, and show that systematic optimization of these components leads to substantial improvements over contemporary non-LLM baselines.

\subsection{Our Baseline Models}

From the related work we chose two NL2SQL models as baselines for our experiments: RASAT \citep{qi-etal-2022-rasat} and SmBoP \citep{rubin-berant-2021-smbop}. Our choice was based on the fact they are both light-weight compared to today's LLMs, but are also very different from each other, both in model architecture and size, which could reveal interesting interactions with the studied extensions.

\textbf{RASAT} utilises the T5 architecture \citep{raffel2020exploring}: it is a transformer sequence to sequence model augmented with relation-aware self-attention in the encoder that can leverage a variety of relational structures, while inheriting the pretrained parameters from the T5 model effectively. It can incorporate relations such as schema encoding, schema linking and syntactic dependency of the question into a unified relation representation.

\textbf{SmBoP} uses GraPPa embeddings \citep{yu2020grappa}, a RAT-SQL encoder \citep{wang-etal-2020-rat} and a novel decoder that uses relational algebra as an intermediate representation. It is a semi-autoregressive bottom-up parser that constructs at decoding step $t$ the top-K sub-trees of height $\le t$. Compared to top-down autoregressive parsing, the bottom-up SmBoP is more efficient as it can decode all sub-trees of a certain height in parallel. Additionally, it learns representations for meaningful semantic sub-programs at each step, allowing it to always generate valid and executable SQL. 

In contrast, RASAT has an unconstrained search space at the decoding step and thus often produces invalid expressions. To mitigate this RASAT uses a filtering or reranking mechanism (PICARD \cite{scholak-etal-2021-picard}) which helps ensure that it generates valid SQL. However, as we are developing our own Reranker module we remove PICARD from the architecture and only compare with the baseline RASAT model.

\section{Component Integration and Evaluation} \label{s:intrinsic}

We develop multiple model architectures by treating the different extensions as modular components that can be added to, or removed from, a baseline NL2SQL system. We use two baselines for our experiments---SmBoP 
and RASAT---
and integrate each component individually, as well as combine them where possible, resulting in hundreds of component combinations\footnote{See Appendix \ref{a:all_experiments} for a comprehensive listing of results obtained for the different baseline and component combinations.}. In this section (\S\ref{s:intrinsic}) we briefly present the integration and evaluation of three of the components: the NatSQL intermediate representation, the GAZP data synthesis process, and a NL2SQL token preprocessing method. We use these systems as building blocks to develop a novel reranker model, which we present in \S\ref{s:reranker}. Finally in \S\ref{s:extrinsic} we present an extrinsic evaluation of the reranker models in combination with the system components from \S\ref{s:intrinsic}, as well as a Shapley analysis used to identify the best performing components.

We use the Spider NL2SQL dataset \cite{yu-etal-2018-spider} as the core dataset in all our experiments and our system evaluations are performed using the Spider evaluation script\footnote{\url{https://github.com/taoyds/spider/blob/master/evaluation.py}}. The script calculates two evaluation metrics to evaluate SQL outputs: Exact Matching (EM) and EXecution accuracy (EX). EM measures whether the predicted query as a whole is equivalent to the gold query and is evaluated as correct only if all of its components are correct. 
EX can be considered an alternative metric as all the databases in Spider have executable SQLite files, meaning we can execute the queries and compare the outputs, rather than the queries themselves. We present both EM and EX in our evaluation results.

\subsection{NatSQL Intermediate Representation}

We use the NatSQL intermediate representation \citep{gan-etal-2021-natural-sql} with both SmBoP and RASAT, neither of which has to our knowledge been attempted in the literature. 
In order to train the vanilla RASAT model to predict the query in NatSQL format, for each of the queries in the Spider train set we used NatSQL queries provided by Gan et al. as the target labels for training the RASAT model. During the inference step we first predict the NatSQL query from the NLQ by using the trained RASAT model and then convert it to SQL format by using the provided conversion script\footnote{\url{https://github.com/ygan/NatSQL}}. 

However, integrating NatSQL with SmBoP presented a significant challenge. Given that SmBoP uses the SQL grammar to generate SQL queries and also parses the target SQL queries to calculate the loss function while training, combining NatSQL with SmBoP required writing the NatSQL grammar rules in the required format and developing a NatSQL parser using this grammar. We also explored different ways of integrating NatSQL with SmBoP and developed three versions of modified SmBoP models, which we call NatSmBoP. Details on this development work are included in Appendix \ref{a:natsmbop}.

The literature indicates that models that leverage NatSQL consistently show performance improvements, likely due to the fact that NatSQL is a simpler representation than the original SQL and is thus easier to learn and predict. We would thus expect to observe the same behaviour when combining NatSQL with RASAT and SmBoP. Our results show that using NatSQL with RASAT significantly improves performance (up by 0.9\% EX and 1\% EM), but surprisingly using it with SmBoP significantly decreases performance compared to its vanilla baseline (down by 3.7\% EX and 1.7\% EM; see full results table in Appendix \ref{a:all_experiments}). 

The improvements from combining RASAT with NatSQL show it can be a valuable addition to the right model. In an effort to streamline the paper and somewhat reduce the number of possible evaluations, we 
perform further experiments only on the best systems established above, namely vanilla SmBoP and RASAT with integrated NatSQL, which we henceforth refer to as NatRASAT.

\subsection{GAZP Synthetic Data Fine-Tuning}

We implement an example of synthetic data generation called Grounded Adaptation for Zero-shot Executable Semantic Parsing (GAZP) \cite{zhong2020grounded}. It is able to generate in-domain synthetic NLQ-SQL pairs based only on a given database and schema. To generate the synthetic data for Spider we first follow the steps from \citet{zhang2019editing} to preprocess SQL logical forms and use the GAZP code as provided by the authors\footnote{\url{https://github.com/vzhong/gazp}}.

We used GAZP to generate 40,875 synthetic NLQ-SQL pairs, however we found that 18,458 were duplicates (45.16\%), and the duplication rate increased dramatically the more samples were generated. Removing duplicates to filter out redundant samples is a direct way to improve on the quality of generated synthetic data \cite{lee-etal-2022-deduplicating,almazrouei2023falcon}, so we removed all duplicated entries. In order to maximise the benefit from the remaining synthetic data we 
explored two variables: training on different synthetic dataset sizes and different distributions of SQL difficulty. After some experimentation we found that using the largest possible amount synthetic data provides the best performance (data in Appendix \ref{a:gazp}). In our evaluations we thus use all our deduplicated GAZP data, i.e. 22,417 unique synthetic samples.

We then integrate the synthetic GAZP data into the model by first training a baseline model on all available deduplicated synthetic data, then fine-tuning that model on the human-generated gold-standard Spider data. This allowed us to train a stronger baseline on the somewhat noisy synthetic samples and then steer further inference in the right direction by fine-tuning on the gold-standard Spider data. We observe significant improvements over both vanilla SmBoP (up by 1.7\% EX and 1.5\% EM) and over NatRASAT (up by 2.5\% EX and 2.3\% EM; see full results in Appendix \ref{a:all_experiments}).

\subsection{Preprocessing}

The final NL2SQL extension we examine is a SotA token preprocessing method (PP) introduced by \cite{rai-etal-2023-improving}, aimed at inducing a more natural tokenisation which is shown to consistently improve model performance. To achieve this, they employ two techniques:  
at the token level their method preserves the semantic boundaries of tokens produced by LM tokenisers (e.g. by tokenising \textit{``pet\_age''} as \textit{``pet'', ``\_''}, and \textit{``age''}, rather than \textit{``pe''}, \textit{``t\_a''}, and \textit{``ge''}), while at the sequence level they propose using special tokens to mark the boundaries of components aligned between input and output. Broadly, this involves adding white spaces and handling naming conventions in database schema and SQL queries. 

We use their token preprocessing implementation as provided\footnote{\url{https://github.com/Dakingrai/ood-generalization-semantic-boundary-techniques}} and apply it to the input SQL queries and database schema before training. We have found that for the SmBoP system applying preprocessing slightly reduces performance (down by 0.5\% EX and EM), while applying it to NatRASAT's pipeline improves scores by 2.1\% EX and 2.7\% EM (see Appendix \ref{a:all_experiments} for more details).

\section{Novel Final Beam Reranker}
\label{s:reranker}

\subsection{Reranker Training Data}
\label{ss:gain_chart}

The final component we study is a final beam Reranker. However, to develop such a model from scratch we require training data in the form of final beams. To provide these, we identify the best performing system combinations developed so far: with SmBoP we observe the best performing combination as SmBoP+GAZP (without preprocessing or NatSQL) on  both EX (76.7\%) and EM (76.7\%). Meanwhile, NatRASAT does best on EX when combined with GAZP (76\%), but does best on EM when combined with Preprocessing (76.2\%). However, in the context of developing a Reranker it is not sufficient to only look at performance scores, but we also need to consider performance potential in the final beam: given a model, how many of its final beams contain the correct query at position in the beam, and in which position? 

\begin{figure}[h!]
	\centering
        \includegraphics[width=1\columnwidth]{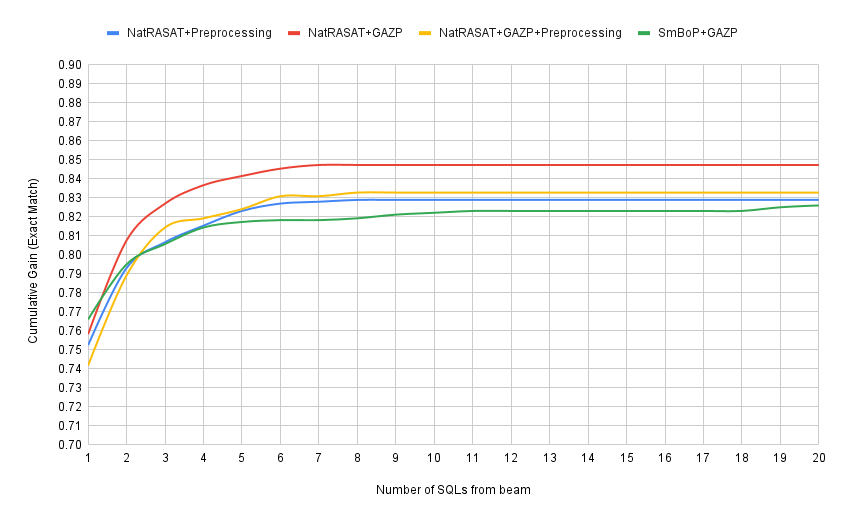}
	\caption{Cumulative gain plot depicting the relationship between beam size and candidate accuracy for the SmBoP and RASAT systems.
    \label{f:beam_coverage-smbop-rasat}}
\end{figure}

Inspired by the work of \citet{klubicka-etal-2018-worth}, we frame this as a gain analysis: Figure \ref{f:beam_coverage-smbop-rasat} presents a cumulative gain chart, which depicts the relationship between beam size (x-axis) and candidate accuracy (y-axis) for the best SmBoP and three NatRASAT systems. The plot lines show how performance improves with beam size, as it is more likely to contain the correct SQL. This analysis clearly shows that NatRASAT+GAZP has the highest performance potential, so we choose that model for our Reranker development. In order to study the effects of training a reranker on a different model's beams than the one it is applied to, in parallel we also use the beams from SmBoP+GAZP.  

\subsection{Reranker Development}

Our reranker work is conceptually inspired by the N-best Reranker paper \citep{zeng2023n}, which introduces a reranking approach combining a query plan classifier and schema linking heuristics that are applied at inference during beam search and serve as a guide for the model to choose valid SQL queries that are more likely to be correct, thus significantly improving performance. However the implementation is not publicly available and is impossible to reproduce based on the paper alone.

Rather than train a Query Plan to guide the reranking, we develop a set of models that directly rerank an already generated ranked beam of candidate SQLs. Here we have no interest in queries other than the best candidate, so the objective of each of these models is to select the best candidate SQL if it is present in the final beam. For this purpose developed two different types of models: \textit{SQL Verifier} and \textit{SQL Selector}\footnote{We will publish the Reranker code and models on GitHub upon publication}. As these are model agnostic, they can then be applied as Rerankers on top of any NL2SQL model, by taking the beam generated by the model as input and selecting the best candidate SQL from this beam. We first describe the SQL Verifier and SQL Selector, then discuss how they can be used as Rerankers in \S\ref{ss:reranking}.

\subsection{SQL Verifier}

The SQL Verifier model is designed to automatically verify whether a candidate SQL is correct for a given NLQ. We finetuned a RoBERTa-based \citep{liu2019roberta} sequence classifier to predict this by using as input the candidate SQL and NLQ and serialised schema information, by concatenating all three into a single input sequence. To train the SQL Verifier model we selected the first 5 candidate SQL queries from the beam of each sample in the Spider train dataset generated by our best-performing NL2SQL models (as established in \S\ref{ss:gain_chart}). Then the trained SQL Verifier model was evaluated intrinsically using a dataset consisting of the first 5 candidate SQL queries from the beam of each sample in the Spider dev set generated by the same best-performing model. 
We ensure that the gold standard SQL of each sample is in the dataset by inserting it if it is not already in the first 5 positions of the beam. This ensures that there will be at least one correct SQL in each beam, though each beam can contain more than one correct SQL apart from the gold standard. Consequently, the label distribution of positive (correct SQL) to negative (incorrect SQL) is not guaranteed to be 20\%:80\% in either the training or validation datasets. The actual label distribution in the training and validation datasets is shown in Table \ref{tab:verifier-labeldistribution}. We observe that both positive (correct SQL) and negative (incorrect SQL) samples have a balanced label distribution, and this is consistent within the training and validation set. The SQL Verifier model is a binary classification model and therefore we measured Precision, Recall, and the F1 score of the models trained on beams generated by our best RASAT-based model and best SmBoP-based model, shown in Table \ref{tab:verifier-intrinsic}. In order to analyse the usefulness of schema information for SQL verification we conducted an ablation study by training models without the schema information, and include results in the last 3 columns of Table \ref{tab:verifier-intrinsic}. These intrinsic evaluations indicate that the impact of schema information may not be significant, as SmBoP+GAZP performs better without it, while NatRASAT+GAZP performs comparably.

\begin{table*}[h]
\centering
\begin{tabular}{l|p{0.5in}p{0.5in}p{0.75in}|p{0.5in}p{0.5in}p{0.75in}}
\hline
\multirow{3}{*}{\shortstack[l]{Model used to \\ generate datasets}} & \multicolumn{3}{c|}{Training dataset} & \multicolumn{3}{c}{Validation dataset}  \\ 
\cline{2-7}						
	&	Positive samples	&	Negative samples	&	\% of positive samples	&	Positive samples	&	Negative samples	&	\% of positive samples	\\ \hline
NatRASAT+GAZP & 14258 &	20276 &	41.29\%	& 2172 & 2905 & 42.78\%	\\
SmBoP+GAZP & 15174 & 19826 & 43.35\% & 2196 & 2974 & 42.48\%	\\	
\hline
\end{tabular}
\caption{Label distribution of training and validation dataset used for SQL Verifier models}
\label{tab:verifier-labeldistribution}
\end{table*}

\begin{table*}[h]
\centering
\begin{tabular}{l|ccc|ccc}
\hline
\multirow{2}{*}{\shortstack[l]{Model used to \\ generate datasets}} & \multicolumn{3}{c|}{With Schema information} & \multicolumn{3}{c}{Without Schema information}  \\ 
\cline{2-7}										
	&	Precision	&	Recall	&	F1	&	Precision	&	Recall	&	F1	\\ \hline	
NatRASAT+GAZP	&	0.7897	&	0.9613	&	0.8671	&	0.7754	&	0.9710	&	0.8622	\\
SmBoP+GAZP	&	0.7756	&	0.9663	&	0.8605	&	0.7897	&	0.9731	&	0.8719	\\ \hline	
\end{tabular}
\caption{Intrinsic evaluation results of SQL Verifier models}
\label{tab:verifier-intrinsic}
\end{table*}

\subsection{SQL Selector}

We developed two categories of SQL Selectors, which directly select the best candidate SQL query from a given list of candidates. One category selects the best SQL query from a list of 5 candidate SQLs, framed as a classification problem which uses as input the NLQ, serialised schema, and the candidate SQLs. We used GraPPa \citep{yu2020grappa} to encode the inputs and a linear layer for the classification. To train this model, we ensure that there exists the correct SQL in the list of 5 candidates by inserting the gold standard SQL into the first position if the list does not contain it.
We also maintain the distribution of the positions of correct SQLs in the training set (beams from the Spider training set) and validation set (beams from the Spider development set) as same. To maintain this we randomly shuffle the candidate SQLs in the beams (\textit{Random} approach). However, this ignores the model's ability to place correct SQLs in first positions. So in an alternate approach we maintained the observed geometric distribution of the position of correct SQL in the validation set by applying a heuristic\footnote{If the correct SQL is in the first position of the beam we maintain the beam as it is with the probability of 0.7. With probability 0.3, we swap that correct SQL from the first position with possibly incorrect SQL from the second position and continue this process recursively on the remaining part of the beam, excluding the first position.} and reordered each beam from the training set to ensure that it also follows the same distribution observed in the evaluation set (\textit{Geometric} approach).

The second category is a Binary SQL Selector in which the model will select the best SQL from a list of 2 candidates. Similar to the previous category of SQL Selectors, we select the first 2 candidate queries from the beam of each sample in the Spider train set generated by the best-performing model to train the binary classification model. To ensure that the correct SQL is in the training set we insert the gold standard SQL from spider data into the first position if not already present. We maintain the distribution of correct queries in the training set as proposed in the previous paragraph. 

We used the best RASAT-based model and the best SmBoP-based model to generate training and validation datasets for SQL Selector models from Spider train and dev sets. The SQL Selector is a classification model and therefore we used accuracy as the evaluation metric. We observed that geometrically distributed beams consistently outperform randomly shuffled beams and binary selection results in a better validation accuracy (See Appendix \ref{a:selector} for the results).

\subsection{Reranking}
\label{ss:reranking}

We use the SQL Verifier models and SQL Selector models to develop three kinds of Rerankers: 

\paragraph{SQL Verifier-based Reranking} This reranker filters the input beam by using an SQL Verifier model and then selecting the first candidate from the filtered rank list as the output. For the filtering each candidate SQL in the beam is verified by using the SQL Verifier model and filtering out those SQL queries which are not verified by the model. If the filtered ranked list is empty, then select the first candidate from the original ranked list.

\paragraph{SQL Selector-based Reranking} This category of reranker selects the best SQL query from the first 5 candidate SQL queries in the input beam by using trained SQL Selector models.

\paragraph{Hybrid Reranking (Binary SQL Selector)} This reranker combines both SQL Verifier models and SQL Selector models. The performance of existing natural language to SQL conversion models is generally good (above 70\% accuracy in most cases) and therefore in most cases (around 70\% of samples) the first SQL candidate in the generated beam will be the correct SQL. By exploiting this, we created binary beams--beams with exactly two candidate SQLs for a sample, with the help of trained SQL Verfier models. In these binary beams, the first candidate SQL from the input beam is directly selected as the first candidate SQL. Then to select the second candidate we filtered the remaining SQL queries in the input beam (excluding the first candidate) by using SQL Verifier models and selected the best from the filtered ranked list. Then we used Binary SQL Selctor models to select the best candidate SQL from these binary beams.

In \S\ref{s:reranker} we have discussed only intrinsic evaluations of the SQL Selector and Verifier (and present more in Appendix \ref{a:selector}), while we discuss their extrinsic evaluations applied as Rerankers in \S\ref{s:extrinsic}. 

\section{Extrinsic Evaluation in a NL2SQL Architecture}
\label{s:extrinsic}

After developing and implementing all the individual components---NatSQL, GAZP, Preprocessing and Rerankers---we combine them to identify the best performing combination of components and study their interactions. This yields hundreds of results which are impossible to fit and discuss in detail in the main body of the paper---instead, here we only summarise the results, presented Table \ref{t:summary}, and discuss some more interesting cases and perform additional analysis, while we include the full results in Appendix \ref{a:all_experiments}.

\begin{table}[htbp]
    \centering
    \begin{tabular}{l|c|c}
     &	\textbf{EX}	&	\textbf{EM}	\\ \hline
    SmBoP & 0.750 & 0.749 \\
    RASAT & 0.727 & 0.727 \\ 
    NatSmBoP & 0.712 & 0.723 \\
    NatRASAT & 0.736 & 0.735 \\ \hline
    SmBoP+GAZP & 0.767 & 0.765 \\
    NatRASAT+GAZP & 0.760 & 0.758 \\
    SmBoP+PP & 0.745 & 0.745 \\
    NatRASAT+PP & 0.757 & 0.762 \\
    SmBoP+GAZP+PP & 0.754 & 0.754 \\
    NatRASAT+GAZP+PP  & 0.744 & 0.742 \\ \hline
    NatRASAT+PP+RR1 (A)& \textbf{0.772} & 0.766	\\	
    NatRASAT+GAZP+RR2 (B)&	0.767	&	\textbf{0.771}	\\ \hline	
    \end{tabular}
    \caption{Summary of full evaluation results on Spider dev set. RR1 is the reranker trained on data from SmBoP+GAZP model and RR2 is the reranker trained on data from NatRASAT+GAZP model}
    \label{t:summary}
\end{table}

Table \ref{t:summary} shows that out of all our combined systems, there are two that compete for first place: (A) RASAT + NatSQL + Preprocessing + Schema-based SQL Verifier (trained on data from SmBoP+GAZP model); and (B) RASAT + NatSQL + GAZP + Schema-based SQL Verifier (trained on data from NatRASAT+GAZP model). System A is the overall winner on EX (up 4.5\% over vanilla RASAT), while System B is the overall winner on EM (up 4.4\% over vanilla RASAT). It can be said that System A in some capacity uses each individual component that we analysed: while it is not directly based on SmBoP or GAZP, its Reranker was trained on data provided by the SmBoP+GAZP system. However the architecture itself does not explicitly integrate these components; similarly, system B does not integrate SmBoP or Preprocessing, indicating that even the best do not necessarily derive their performance from a combination of all pipeline optimisation approaches.

\begin{table*}[htbp]
    \centering
    \begin{tabular}{l|cc|cc|cc|cc}
\hline
	&	\multicolumn{2}{c|}{NatSQL}	&	\multicolumn{2}{c|}{GAZP}	&	\multicolumn{2}{c|}{PP}	&	\multicolumn{2}{c}{Reranker}\\
	&	EX	&	EM	&	EX	&	EM	&	EX	&	EM	&	EX	&	EM	\\
 \hline
SmBoP + SmBoP-Reranker	&	-3.150	&	-1.775	&	0.900	&	1.425	&	-0.483	&	-0.333	&	-0.483	&	-0.533	\\
SmBoP + RASAT-Reranker	&	-4.425	&	-2.875	&	1.250	&	1.875	&	-0.517	&	-0.433	&	-0.750	&	-0.867	\\
\textbf{SmBoP Overall}	&	{\color{red}-3.788}	&	{\color{red}-2.325}	&	{\color{blue}1.075}	&	{\color{blue}1.650}	&	-0.500	&	-0.383	&	-0.617	&	-0.700	\\
\hline
RASAT + SmBoP-Reranker	&	0.900	&	1.000	&	-0.075	&	0.000	&	0.175	&	0.400	&	0.825	&	0.750	\\
RASAT + RASAT-Reranker	&	0.900	&	1.000	&	0.425	&	0.450	&	0.075	&	0.350	&	0.975	&	1.200	\\
\textbf{RASAT Overall}	&	{\color{blue}0.900}	&	{\color{blue}1.000}	&	0.175	&	0.225	&	0.125	&	0.375	&	{\color{blue}0.900}	&	{\color{blue}0.975}	\\
\hline
    \end{tabular}
    \caption{Shapley values of different components. A system's most positive contributors are shaded blue, while most negative contributors are shaded red.}

    \label{t:shapley-combined}
\end{table*}

\subsection{Shapley Value Analysis}

However, if we wished to answer the question of which of the extensions contribute most to the system's performance, simply comparing the different models' performance scores does not offer a straightforward answer regarding individual extension interactions. To obtain more insight, we measured the usefulness of NatSQL, Preprocessing, GAZP, and Reranker by adopting the concept of Shapley value from cooperative game theory \citep{shapley1953}. Here we treat the extended NL2SQL system as an n-player cooperative game and each components as a player. Then the Shapley value of a component is a number describing the contribution of that component to the performance of the extended system. The higher the Shapley value of a component the greater the contribution the component made to the system's performance. We separately calculated Shapley values of each of these extensions on four different system frameworks, i.e., SmBoP-based and RASAT-based systems with SmBoP-based and RASAT-based Rerankers. We also calculated the overall Shapley values on both SmBoP-based and RASAT-based systems by treating SmBoP-based and RASAT-based Rerankers as two independent components\footnote{Note that, to calculate exact Shapley values we require the scores of all model combinations. But all such scores were not available and therefore we calculated the approximate Shapley values by using the available scores.}.

The obtained Shapley values (measures of usefulness) are presented in Table \ref{t:shapley-combined}, which shows that GAZP is the only extension that usefully combines with any of SMBoP-based systems, all other extensions have negative Shapley values. By comparison, all the extensions improve the performance of the RASAT-based model (positive Overall Shapley values) and of the four extensions NatSQL and the Reranker are significantly more beneficial than GAZP and Preprocessing.

\subsection{Discussion}

While our best systems offer improvements upwards of 4.5\% over their vanilla baselines, indicating positive interactions between the studied components, we have also observed a surprising amount of negative interactions. We discuss some of these failure cases below, and find that that a ``combine everything'' approach may not be a good rule of thumb when optimising NL2SQL systems. 

\paragraph{Preprocessing and GAZP} It is intriguing that the positive effects of combining Preprocessing and GAZP fine-tuning do not compound: while applying either of them to RASAT individually improves performance, applying both together underperforms the individual ones (but outperforms vanilla baseline). On SmBoP this result is similar, except the preprocessing on its own underperforms when compared to baseline, while GAZP outperforms baseline. Combining them outperforms baseline slightly, but significantly underperforms when compared to SmBoP+GAZP, indicating the two components do not play well together. We are puzzled as to why this is, given that the Preprocessing approach is quite surface-level, while the generated synthetic data is in the same format as gold standard data and should not be affected by the changes in tokenisation. 

\paragraph{NatSmBoP} Although the integration of NatSQL with SmBoP required a significant amount of work, and the addition of preprocessing and a Reranker improves the performance of NatSmBoP, the performance of any NatSmBoP combination is well below that of vanilla SmBoP. We found this to be unexpected, as both the literature and our other experiments have shown that generally adding NatSQL to other systems significantly improves performance, robustness and generalisability. We performed a beam coverage analysis on NatSmBoP (see Appendix \ref{a:natsmbop}, Figure \ref{f:beam_coverage}) and found that all three variants dramatically outperform vanilla SmBoP in their gain charts. In other words, the right reranker could improve their performance, however none of our Rerankers were able to leverage this to the point of outperforming vanilla SmBoP.

Given that SmBoP already uses relational algebra as an intermediate representation, we wonder whether perhaps there is some kind of inherent incompatibility between the relational algebra and NatSQL intermediate representations, which might cause NatSmBoP to consistently underperform.

\paragraph{SmBoP} It seems that overall SmBoP is quite an inflexible system. The majority of our approaches involving SmBoP led to no improvements over the basline, or even decreased the accuracy scores. The only thing that seemed to help is the fine-tuning with GAZP, which increased performance overall, but everything else---combining with NatSQL, adding preprocessing or adding any Reranker, or even doing them together---had a very limited or negative effect. We suspect that SmBoP is not very adaptable and is not compatible with significant changes to its framework or pipeline. 

\paragraph{Reranking} It is interesting that our best performing models rely on an SQL Verifier with Schema Information---recall that the intrinsic evaluation results in Table \ref{tab:verifier-intrinsic} have shown that the impact of schema information is not significant. However the extrinsic evaluation, i.e. when applied as a reranker at the end of a full NL2SQL pipeline, shows that this is in fact the best-performing reranker model. It is likely that intrinsically the schema information does not help much, however when the system is much more complex the addition of that information seems to help the model make better decisions.

In any case, the different new flavours of our Reranker have an undeniably positive impact when combined with our systems, improving performance across the board and showing strong improvements when compared to systems that do not use a reranker at all. In fact, our NatRASAT+Reranker model even outperforms the original RASAT+PICARD model on both the dev set (higher by 1.8\% EM) and the official Spider test set (higher by 1\% EX\footnote{See system 15 on the Spider leaderboard: \url{https://yale-lily.github.io/spider}}), indicating that a good reranker is a crucial component in improving NL2SQL system performance. 

It is also important to note that both the Verifier and Selector are model agnostic---each only takes a beam of SQL candidates as input and makes a prediction, meaning they can be integrated with any underlying NL2SQL model. This also allows integration with large language models, provided they are prompted to output a beam of SQL candidates. We briefly explored this option (see Appendix \ref{a:llama}) but found the results limited, while a more thorough investigation was out of scope for this paper.

\section{Conclusion}

With the goal of raising the performance ceiling of more lightweight NL2SQL systems, we have identified a number of components in state of the art approaches that can be merged together to form novel combinations, while still retaining a relatively small memory footprint. The best performing system increases both EX and EM accuracy scores significantly above vanilla RASAT and rivals the best non-LLM systems on the Spider leaderboard. More importantly, our work has shown that applying all model extensions together is not the most helpful approach: we have learned that the components' individual benefits do not always compound, and there can be negative interactions. Best results are dependant on interactions between the baseline model and its extensions, as well as between the extensions themselves. Finally, we stress that all the approaches studied here are model agnostic, hence they can be applied to any baseline NL2SQL model, including LLMs, which is a rich area of future work.

\section*{Limitations}

Our main concern when undergoing this study was scope creep. There is actually a considerable amount of work we were forced to leave on the cutting room floor due to the page limit, and even more we could have done but did not have resources for (mainly time). Hence many avenues remain open to research the topics touched upon in the paper.

As an example, we have our choice of baseline models SmBoP and RASAT. While our choice was motivated by the wish to try models that have different architectures (semi-autoregressive with specialised decoder VS auto-regressive LM-bsed) and sizes ($\approx$1GB model file VS $\approx$11GB model file), in hopes of revealing interesting interactions with the other components, this choice is also arbitrary in a way, and we could have used any number of different baselines. On the flip side, this is useful as it leaves many avenues for future work, given that the approaches we have studied are model-agnostic and rely more on manipulating data and outputs, rather than the underlying predictive models themselves.

Furthermore, for the same streamlining reasons we did not have scope to actually combine literally every possible component. Some omissions include NatSmBoP with Preprocessing or GAZP, as well as vanilla RASAT with Preprocessing or GAZP. These were decisions made purposefully to guide the discussion and decision-making process in a cascaded fashion, as it is unlikely the the lowest performing systems would outperform the stronger baselines when combined with these components. However these are only assumptions and we do not present evidence to experimentally verify this.

Another example is our choice of the training and evaluation dataset. We only considered Spider, which only contains English natural language queries (a more general limitation of the work), and contains a constrained number of SQL types and examples of syntax. While Spider strives to be generalisable, it does not contain examples of commonly used keywords such as COMPARISON, TREND or TIME\_PERIOD. Supplementing training with a broader set of queries, or even just evaluating our models on other benchmarks (such as WikiSQL \cite{zhong2017seq2sql}, SParC \cite{yu-etal-2019-sparc} or CoSQL \cite{yu-etal-2019-cosql}), would improve their generalisability. However, our main goal here was to perform and present a comprehensive, systematic analysis of architecture component combinations and their interactions, and given the density of our work at present, including even one additional dataset would require us to produce an additional 351 results, exponentially increasing the complexity of the work, research time and workload, as well as energy footprint. 

While we acknowledge a number of the work's limitations, we stress that all our choices have been made in a sound, informed and methodologically consistent manner. Here we simply highlight just how many choices have been made along the way, and how quickly the number of alternative paths grows the further back up the decision tree we look. While we believe that the work is fundamentally solid, each choice could have made for a drastically different suite of experiments and could potentially have yielded different results. In fact, we find this to be a very exciting motivator for future work.

\section*{Acknowledgements}

This research was conducted with the financial support of Research Ireland under Grant Agreement No. 13/RC/2106\_P2 at the ADAPT Research Ireland Research Centre. ADAPT, the Research Ireland Research Centre for AI-Driven Digital Content Technology, is funded by Research Ireland through the Research Ireland Research Centres Programme, and is co-funded under the European Regional Development Fund. In addition, this research was funded by and developed in collaboration with Huawei Ireland.

\bibliographystyle{plainnat}  
\bibliography{references}

\appendix

\section{NatSmBoP Parser}\label{a:natsmbop}

We developed a parser to generate the parsed tree from a query in NatSQL format. As the first step for developing a parser, we listed all grammar rules of NatSQL based on the grammar mentioned in the original paper \citep{gan-etal-2021-natural-sql}. We then converted this original grammar to Backus-Naur Form (BNF) such that the expression of each rule contains sequences of two non-terminal symbols or one terminal. This is to ensure that the parsed tree will always be a binary tree, which is necessary in order to incorporate NatSQL with SmBoP model, because the SmBoP model is compatible with binary parsed trees only. Note that the SmBoP decoder selects at most two subtrees to combine and generate another parsed subtree in each step of decoding and therefore the model will always generate a binary parsed tree. We also observed that the NatSQL grammar mentioned in the paper is not complete to parse every NatSQL query of the Spider data and therefore we modified the grammar further by adding some extra rules to make it complete, at least for all NatSQL queries in the Spider data. All the grammar rules of NatSQL after the above-mentioned modifications are shown in Table \ref{t:natsql} and \ref{t:natsql-terminal}. We developed a NatSQL parser by using these grammar rules to convert NatSQL queries to parsed binary tree format. We used the modgrammar package\footnote{\url{https://pypi.org/project/modgrammar/}} to develop the parser and we verified that all the target NatSQL queries in Spider train set and development set can be parsed by using this parser.

    \begin{table*}[ht]
    \centering
    \begin{tabular}{rcp{11cm}}
        NatSQL & ::= & MainClause WhereGroupOrderByClause | MainClause GroupOrderByClause | MainClause OrderbyClause | MainClause GroupbyClause | MainClause WhereClause | MainClause UNONE \\
        MainClause & ::= & SelectClause FromClause \\
        SelectClause & ::= & SELECT TabCol | SELECT TabCols | SELECT DistinctTabCols | SELECT AggColumn | SELECT Columns \\
        FromClause & ::= & FROM Table \\
        Columns & ::= & TabCol RemColumns | TabCols RemColumns | DistinctTabCols RemColumns | AggColumn RemColumns \\
        RemColumns & ::= & COMMA TabCol | COMMA TabCols | COMMA DistinctTabCols | COMMA AggColumn | COMMA Columns \\
        AggColumn & ::= & AggFunction Operand \\
        AggFunction & ::= & AggFuncName LBRACKET \\
        Operand & ::= & TabCol RBRACKET | TabCols RBRACKET | DistinctTabCols RBRACKET \\
        DistinctTabCols & ::= & DISTINCT TabCol | DISTINCT TabCols \\
        TabCols & ::= & TabCol CombineTabCols \\
        CombineTabCols & ::= & ArithOp TabCol | ArithOp TabCols \\
        WhereGroupOrderByClause & ::= & WhereClause OrderbyClause | WhereClause GroupbyClause | WhereClause GroupOrderByClause \\
        GroupOrderByClause & ::= & GroupbyClause OrderbyClause \\
        WhereClause & ::= & WHERE ConjunctedCondition | WHERE Condition | WHERE Conditions | WHERE WhereCondition \\
        OrderbyClause & ::= & ORDERBY TabCol | ORDERBY TabCols | ORDERBY DistinctTabCols | ORDERBY AggColumn | ORDERBY Columns | ORDERBY OrderBy \\
        GroupbyClause & ::= & GROUPBY TabCol | GROUPBY TabCols | GROUPBY DistinctTabCols | GROUPBY AggColumn | GROUPBY Columns \\
        WhereCondition & ::= & ConjOp Conditions \\
        Conditions & ::= & Condition ConjunctedCondition | Condition ConjunctedConditions \\
        ConjunctedConditions & ::= & ConjunctedCondition ConjunctedCondition | ConjunctedCondition ConjunctedConditions \\
        ConjunctedCondition & ::= & ConjOp Condition \\
        OrderBy & ::= & TabCol, OrderLimit | TabCol Limit | TabCol Order | TabCols OrderLimit | TabCols Limit | TabCols Order | DistinctTabCols OrderLimit | DistinctTabCols Limit | DistinctTabCols Order | AggColumn OrderLimit | AggColumn Limit | AggColumn Order | Columns OrderLimit | Columns Limit | Columns Order \\
        OrderLimit & ::= & Order Limit \\
        Limit & ::= & LIMIT String \\
        Condition & ::= & TabCol CondString | TabCols CondString | DistinctTabCols CondString | AggColumn CondString | AT CondString | TabCol BetweenOp | TabCols BetweenOp | DistinctTabCols BetweenOp | AggColumn BetweenOp | AT BetweenOp \\
        CondString & ::= & CondOp TabCol | CondOp TabCols | CondOp DistinctTabCols | CondOp AggColumn | CondOp String \\
        BetweenOp & ::= & BetweenLHS BetweenRHS \\
        BetweenRHS & ::= & AND String \\
        BetweenLHS & ::= & BETWEEN String\\
    \end{tabular}
    \caption{NatSQL grammar rules}
    \label{t:natsql}
    \end{table*}
    
    \begin{table*}[ht]
    \centering
    \begin{tabular}{rcp{12cm}}
        UNONE & ::= & none \\
        KNONE & ::= & none \\
        SELECT & ::= & "select" \\
        FROM & ::= & "from" \\
        COMMA & ::= & "," \\
        LBRACKET & ::= & "(" \\
        RBRACKET & ::= & ")" \\
        DISTINCT & ::= & "distinct" \\
        WHERE & ::= & "where" \\
        ORDERBY & ::= & "order by" \\
        GROUPBY & ::= & "group by" \\
        LIMIT & ::= & "limit" \\
        AT & ::= & "@" \\
        AND & ::= & "and" \\
        BETWEEN & ::= & "between" \\
        AggFuncName & ::= & "avg" | "count" | "max" | "min" | "sum" \\
        ArithOp & ::= & "-" | "+" \\
        ConjOp & ::= & "and" | "or" | "except" | "except\_" | "intersect" | "intersect\_" | "union" | "union\_" | "sub" \\
        CondOp & ::= & "=" | ">" | "<" | "<=" | ">=" | "!=" | "in" | "like" | "is" | "exists" | "not in" | "not like" | "not between" | "is not" | "join" \\
        Order & ::= & "asc" | "desc" \\
        String & ::= & '0' | '1' | '2' | '3' | '4' | '5' | '6' | '7' | '8' | '9' | 'yes' | 'no' | 'y' | 'n' | 't' | 'f' | 'm' | 'null' | '""' \\
    \end{tabular}
    \caption{NatSQL grammar rules (Terminals)}
    \label{t:natsql-terminal}
    \end{table*}

\subsection{SmBoP with NatSQL}

We modified the existing SmBoP model to generate NatSQL queries from NLQs by using the NatSQL grammar shown in Table \ref{t:natsql} and \ref{t:natsql-terminal}. To train this SmBoP model we generate parsed trees of target queries in NatSQL representation by using the newly developed NatSQL parser. The NatSQL query generated by this modified SmBoP model will be converted to SQL queries by using the publicly available conversion script at the inference time. We developed three different versions of modified SmBoP models called NatSmBoP:

\begin{enumerate}
\item \textbf{NatSmBoP.V.0}: This model is the modified vanilla SmBoP model to incorporate NatSQL grammar. We used the same hyperparameters as in the vanilla SmBoP model except the beam size. In the modified version of SmBoP, there are more tokens in the initial beam and therefore we increased the beam size from 30 to 100.
\item \textbf{NatSmBoP.V.1}: This version is a variant of NatSmBoP.V.0 with a modified loss function. The SmBoP decoder generates parsed tree iteratively in a bottom-up manner. It initialises a beam with all the leaf nodes in the parse tree, in other words, it initialises with a set of most probable parsed (sub)trees with height 1. Then in each iteration, it generates a new beam with a set of most probable (sub)parsed trees with height incremented by 1 from the previously generated beam. So in $i^{th}$ iteration, the beam will have a set of parsed (sub)trees with height $i+1$. After the final iteration, the model selects the best parsed tree from the final beam as the output. The final beam will have exactly one correct parsed tree which is the parsed tree of the target query unless there are multiple possible queries for the question. But in the second last beam there can be more than one correct subtree with a height one less than the height of the correct parsed tree. If the correct parsed tree is a binary tree then there will be exactly two correct parsed trees in the second last beam. This can continue and there can be a higher number of correct subtrees in the corresponding beam when the height of the subtree decreases. The vanilla SmBoP model calculates the loss function from each of these beams with equal weight irrespective of the height of parsed (sub)trees. So there will be more terms from initial beams which contribute to the final loss function than from final beams and this can reduce the impact of loss from the final beam which is more critical for the final prediction. To mitigate this potential issue, in this modified version (NatSmBoP.V.1.) we normalised the loss from each beam with the number of target parsed (sub)trees in that beam. This is similar to the macro averaging of the loss function and therefore we called this new loss function the macro-loss.
\item \textbf{NatSmBoP.V.2}: This version is a variant of NatSmBoP.V.0 with a linear beam size decay. As discussed earlier, the number of target parsed (sub)trees will reduce in each iteration of SmBoP decoding. Therefore the beam size can be reduced in each iteration and that will reduce the number of incorrect (sub)trees. For that, we used a linear beam size decay in which the beam size reduced linearly from the initial beam size of 100 to the final beam size of 10. 
\end{enumerate}   

We report the experimental results of all NatSmBoP variants alongside all other results in Appendix \ref{a:all_experiments}. The results show that none of the the combined NatSQL and SmBoP systems can achieve parity with the vanilla SmBoP system. The best combined system seems to be NatSmBoP.V.0, though it still significantly underperforms vanilla SmBoP on both EX and EM accuracy scores. 

\begin{figure}[h!]
	\centering
        \includegraphics[width=1\columnwidth]{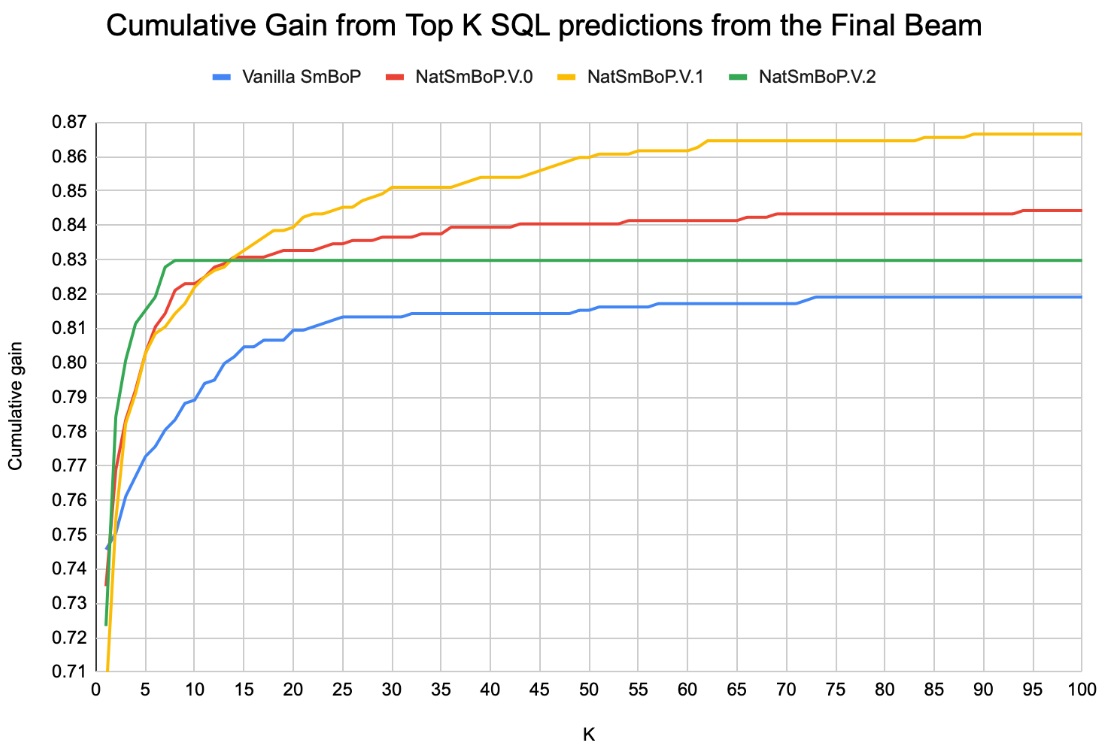}
	\caption{Cumulative gain plot depicting the relationship between beam size and candidate accuracy for the SmBoP and NatSmBoP systems.
    \label{f:beam_coverage}}
\end{figure}

In an effort to better understand whether integrating NatSQL with SmBoP is helpful at all we performed a beam coverage analysis similar to what was done in \S\ref{ss:gain_chart}. It is presented in Figure \ref{f:beam_coverage} as a cumulative gain chart (as inspired by the work of \citep{klubicka-etal-2018-worth}), which depicts the relationship between beam size (x-axis) and candidate accuracy (y-axis) for the SmBoP and NatSmBoP systems. The plot lines thus show how performance can improve the larger the beam size gets, as it is more likely to contain the correct SQL candidate. Our gain chart reveals that although none of the NatSmBoP variants are able to select the correct queries as most likely, they are all able to include them in their final beam more times than vanilla SmBoP. This means that any NatSmBoP variant has a higher performance potential, as it generates more beams with the correct SQL candidate than SmBoP does. This strongly motivates the development and application of a Reranker to this model, as adding an additional model to improve SQL selection could leverage the higher performance potential of NatSmBoP. However as seen in our results, this was not the case so we leave it out of the main discussion.

\section{GAZP Size Experiments}
\label{a:gazp}

We generated a dataset containing 22,417 unique synthetic samples, and during this process monitored the generation of new, unique queries divided by difficulty category. We noticed that by the time we generated around 15,000 unique samples the number of easy queries had already plateaued. Generating more synthetic queries would only generate duplicate easy queries, but was still generating unique medum/hard/extra hard queries. By the time we reached $\approx$22,000 we noted that generation of unique queries of any difficulty slowed down significantly. 

\begin{figure}[h!]
	\centering
	\includegraphics[width=1\columnwidth]{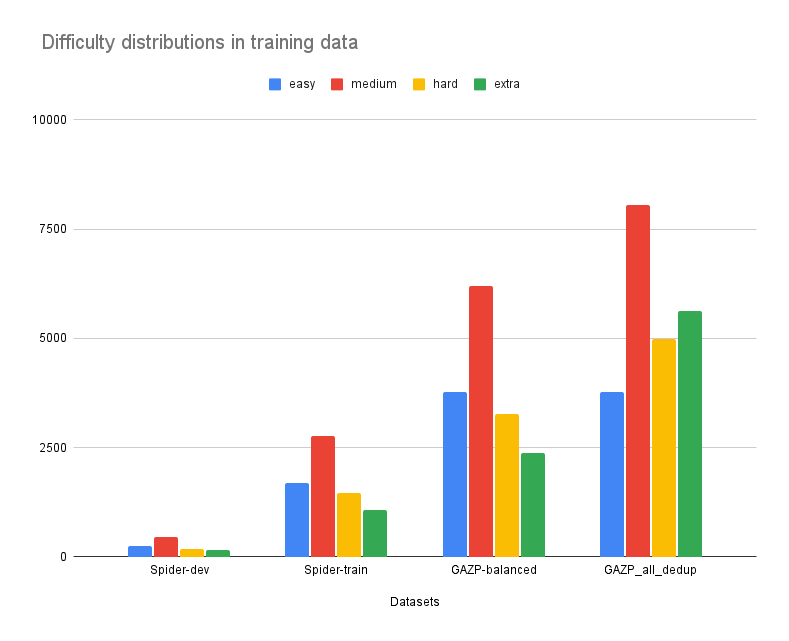}
	\caption{SQL difficulty distributions in Spider dev and train set, as well our largest synthetic GAZP dataset, as well as a Spider-balanced GAZP subset. 
    \label{f:gazp_plot}}
\end{figure}

When using the synthetic data for training, we wondered if it would be more beneficial to emulate the difficulty distribution present in the Spider dev and train sets. We thus trained separate models to examine the impact of training on a synthetic dataset with the exact size and distribution of the Spider train set (7000 samples), the largest possible synthetic dataset that still conforms to Spider's SQL difficulty distribution (15,627 samples) and simply the largest dataset we generated which includes all the deduplicated data (22,417 samples) but does not match Spider's difficulty distribution. Statistics on generated data broken down by SQL difficulty are shown in Table \ref{t:gazp_sizes} and Figure \ref{f:gazp_plot}.

\begin{table*}[htbp]
    \centering
    \begin{tabular}{lcccc}
        \textbf{Difficulty} & \textbf{Spider-dev} & \textbf{Spider-train} & \textbf{GAZP-balanced} & \textbf{GAZP-all-dedup} \\
        \hline
        easy & 248 & 1694 & 3782 & 3782 \\
        medium & 446 & 2777 & 6200 & 8040 \\
        hard & 174 & 1460 & 3260 & 4975 \\
        extra & 166 & 1068 & 2385 & 5629 \\
        \hline
        total & 1034 & 7000 & 15627 & 22417 \\
    \end{tabular}
    \caption{Dataset Statistics}
    \label{t:gazp_sizes}
\end{table*}

The baseline results of different sizes are included in Table \ref{t:gazp_results_full}. We can see that as we include more data, regardless of the SQL difficulty distributions, the scores increase, though the overall performance of the baseline GAZP model is generally in the same ballpark. However, the best approach for integrating GAZP data is to use the maximum available amount of data, regardless of the SQL difficulty distribution. This can be see when comparing the overall EX/EM accuracy scores between GAZP\_7k, GAZP\_15k and GAZP\_22k, where 22k exhibits the best performance. In other words, it is not helpful or important to match the distribution of SQL difficulties present in the Spider train set, as the model performance benefits more from including a larger synthetic sample with an imbalanced distribution of SQL difficulty.

\begin{table*}[htbp]
    \centering
    \begin{tabular}{ll|c|ccc}
        \textbf{Metric} & \textbf{Difficulty} & \textbf{SmBoP} & \textbf{GAZP\_7k} & \textbf{GAZP\_15k} & \textbf{GAZP\_22k} \\
        \hline
        \textbf{EX} & ALL & 0.75 & 0.514 & 0.514 & 0.547 \\
         & easy & 0.87 & 0.613 & 0.613 & 0.629 \\
         & medium & 0.79 & 0.556 & 0.538 & 0.576 \\
         & hard & 0.67 & 0.443 & 0.489 & 0.546 \\
         & extra & 0.54 & 0.325 & 0.325 & 0.349 \\
        \hline
        \textbf{EM} & ALL & 0.75 & 0.505 & 0.527 & 0.546 \\
         & easy & 0.88 & 0.685 & 0.706 & 0.698 \\
         & medium & 0.8 & 0.549 & 0.57 & 0.601 \\
         & hard & 0.64 & 0.368 & 0.414 & 0.431 \\
         & extra & 0.52 & 0.259 & 0.265 & 0.295 \\
    \end{tabular}
    \caption{Results of SmBoP with GAZP-baseline training and GAZP-fine-tuning, compared to vanilla SmBoP.}
    \label{t:gazp_results_full}
\end{table*}

\begin{table*}[h!]
    \centering
    \begin{tabular}{l|c|c|cc}
        \hline
        \textbf{Metric} & \textbf{SmBoP} & \textbf{GAZP\_22k} & \textbf{GAZP\_7k+Spider} & \textbf{GAZP\_22k+Spider\_ft} \\
        \hline
        \textbf{EX} & 0.75 & 0.547 & 0.7 & \textbf{0.767} \\
        \hline
        \textbf{EM} & 0.75 & 0.546 & 0.688 & \textbf{0.765} \\
        \hline

    \end{tabular}
    \caption{Results of SmBoP with GAZP-baseline training and GAZP-fine-tuning, compared to vanilla SmBoP.}
    \label{t:gazp_results}
\end{table*}

Perhaps the more important consideration is how to combine the synthetic GAZP data with the gold-standard Spider data. We identified two possible approaches: (a) simply combining the GAZP and Spider data into a single training dataset\footnote{Note this would require a dataset that is balanced between synthetic and gold-standard data in order for the somewhat noisy synthetic data not to overshadow the contribution of the gold data. Specifically, we combine the 7k Spider samples with 7k GAZP samples for a total train set of 14k samples.}; and (b) first training a baseline model on the synthetic data, then fine-tuning that model on the human-generated gold-standard Spider data. Option (a) streamlines the training implementation, however option (b) allows us to potentially use more synthetic data by training a stronger baseline on the noisy synthetic samples, and then steering further inference in the right direction by fine-tuning. 

We run this intrinsic evaluations on the vanilla SmBoP model (the best baseline model) 
and the results are presented in Table \ref{t:gazp_results}. They indicate that simply combining GAZP and Spider data into a singular training set significantly reduces performance compared to vanilla SmBoP (down by 5\% EX and 6.2\% EM), likely due to the varying quality and inherent noisiness of the synthetic data. The better approach for integrating GAZP data is to use the maximum available amount of data and fine-tune that model on the gold-standard Spider data, which provides a significant performance improvement over the vanilla SmBoP results. Having determined this is the best approach to integrate GAZP, we apply it throughout the evaluation of all our model combinations. 

\section{SQL Selector Evaluation Results}\label{a:selector}
Here we present accuracies of SQL Selector models in Table \ref{tab:selector-intrinsic}.

\begin{table*}[ht]
\centering
\begin{tabular}{l|cc|cc}
\hline
\multirow{2}{*}{\shortstack[l]{Model used to \\ generate training data}} & \multicolumn{2}{c|}{SQL Selector} & \multicolumn{2}{c}{Binary SQL Selector}  \\ \cline{2-5}		
	&	Random	&	Geometric	&	Random	&	Geometric	\\ \hline	
NatRASAT+GAZP	&	0.8346	&	0.8791	&	0.9255	&	0.9255	\\
SmBoP+GAZP	&	0.7979	&	0.8694	&	0.8985	&	0.9197	\\ \hline	

\end{tabular}
\caption{Intrinsic evaluation of SQL Selector models}
\label{tab:selector-intrinsic}
\end{table*}

\section{Reranker with LLMs}
\label{a:llama}

Given the prominence of LLMs we are curious to see whether our Reranker can help when applied in an LLM scenario. Given that the Reranker is model agnostic, in principle it is straightforward to apply it to any model, assuming that the model can provide a ranked list of candidate SQL queries. While we are not able to look directly ``under the hood'' of an LLM in the same way we can in SmBoP or RASAT, we can use a prompt to ask it to generate an ordered list of SQL candidates given a NLQ. We can then apply the reranker to the output of the LLM and assess whether it can help choose the correct SQL from the generated list. 

We approached the evaluation in the following way: as a case study, we used the openly available off-the-shelf model LLaMA, specifically version Llama-2-13b-chat-hf from HuggingFace\footnote{\url{https://huggingface.co/meta-llama/Llama-2-13b-chat-hf}} and set it up as a zero-shot problem, evaluating its performance in three settings: (1) when asked to generate 1 correct SQL translation; (2) when asked to generate a ranked beam of 20 SQL candidates, selecting the 1st candidate in the beam; and (3)
when asked to generate a ranked beam of 20 SQL candidates, using our best-performing Reranker to help select the correct candidate from the beam.

To have LLaMA generate a ranked ``beam'' it not only requires the NLQ, but also the database schema, i.e. table and column information, in order to understand the data structure and be able to provide an accurate prediction. Below is a truncated example of the prompt structure we used to generate outputs: 

\begin{verbatim}
Assume there are multiple possible SQL queries that correctly correspond to a given question. 
Given the following question and database schema, generate the best 20 candidate SQL queries that 
correspond to the question. Format the generated SQL queries as one single python list. 
Ensure that the queries in the list are ordered by most likely to be correct, where the first one 
is most likely. Only respond with one python list and do not output any other text.

Schema:
Table advisor, columns = [*,s_ID,i_ID]
Table classroom, columns = [*,building, room_number,capacity]
...
Foreign_keys = [advisor.s_ID = student.ID, advisor.i_ID = instructor.ID,(...)]

Q: What is the capacity of all the rooms?

\end{verbatim}


% ...



\begin{table*}[h]
    \centering
    \begin{tabular}{l|ccc}
        & \textbf{1 SQL} & \textbf{20 SQLs} & \textbf{20 SQLs RR} \\
        \hline
        \textbf{EX} & 16.60\% & 23.50\% & 23.80\% \\
        \textbf{EM} & 13.70\% & 17.80\% & 18.00\% \\
    \end{tabular}
    \caption{LLaMA performance scores on the Spider dev set.}
    \label{t:llama_eval}
\end{table*}

The results for each case are presented in Table \ref{t:llama_eval}, where it seems that LLaMA is not very good at NL2SQL in a zero-shot scenario. While using our Reranker does help, the overall improvement on the dev set is minimal. It is intriguing however that LLaMA's accuracy is somewhat higher when generating beams of 20 candidates, versus asking it to generate only 1 candidate. To better understand what's going on when applying the reranker to LLaMA, we also examined the confusion matrix and counted the cases where the reranker helped LLaMA and where it did not, presented in Table \ref{t:confusion}.

\begin{table*}[h]
    \centering
    \begin{tabular}{l|ccc}
        & \textbf{llama = RR} & \textbf{RR+} & \textbf{RR-} \\
        \hline
        \textbf{1st SQL correct} & 183 & 0 & 1 \\
        \textbf{1st SQL incorrect} & 723 & 3 & 124 \\
    \end{tabular}
    \caption{Confusion matrix of Reranker predictions on LLaMA outputs.}
    \label{t:confusion}
\end{table*}

The first row shows total cases where the 1st SQL candidate was correct, and second row shows cases where 1st SQL was incorrect. When applying the Reranker to the beam, the first column shows the number of cases where the Reranker selected the 1st SQL candidate (i.e. made the same prediction as LLaMA). The second column shows number of cases that the Reranker chose a different SQL from the beam (i.e. not the 1st candidate) and chose correctly, while the third column shows number of cases that the Reranker chose a different SQL from the beam and chose incorrectly. 
We're most interested in row 2, which shows that the Reranker chose the correct SQL in 3 cases (thus improving performance). It is also interesting that in 124 cases the Reranker identified that the 1st SQL was incorrect and chose a different candidate from the beam, but still chose an incorrect one. However, this is due to the fact that none of the other queries in the beam were correct, as only 3\% of the samples in LLaMA's output contain the correct SQL in the beam (not in 1st place). 

So while overall the Reranker has a small positive impact on LLaMA's performance, the biggest problem is that if LLaMA does not generate the correct candidate in the first position, then it is likely not going to generate it anywhere else in the beam. This means the Reranker really does not have much opportunity to help in this case, as LLaMA's performance is generally very low.

It is worth mentioning that engineering the prompts for LLaMA took quite some time and effort. Our prompt structure was partially inspired by prompts developed by \cite{pourreza2023dinsql}, though they used a few-shot approach on GPT-4 and achieved much better performance in their setup. It is possible that our prompt structure does not interact with LLaMA very well when trying to tease out NL2SQL knowledge. 
Even still, a lot of effort went into not just getting LLaMA to consistently generate a list of 20 candidate SQLs, but also to get it to output this in a particular format. The outputs were quite inconsistent and even in the best cases required different post-processing approaches. If in future work we wanted to improve LLaMA's performance on the task, we suspect more comprehensive prompt-engineering combined with a few-shot approach are needed to get higher quality and more consistent outputs overall.

\section{Full Cohort of System Combinations Experiments}
\label{a:all_experiments}

Here we present the full cohort of experimental results for our ablation experiments combining the model peripherals studied in this research. The results are found at the end of the paper, in Tables \ref{tab:ex-smbop-reranker}, \ref{tab:ex-rasat-reranker}, \ref{tab:em-smbop-reranker} and \ref{tab:em-rasat-reranker}.

\begin{table*}[h]
\centering
\scalebox{0.83}{
\centering
\begin{tabular}{|l|c|cc|cc|cc|}
\hline
&	\multirow{2}{*}{Baseline}	&	\multicolumn{2}{c|}{SQL Verifier} &	\multicolumn{2}{c|}{SQL Selector} &	\multicolumn{2}{c|}{Binary SQL Selector} \\
	&		& No Schema & \textbf{Schema} & Geometric & Random & Geometric & Random \\ \hline

SmBoP	&	0.750	&	0.751	&	0.747	&	0.733	&	0.669	&	0.722	&	0.726	\\
SmBoP+Preprocessing	&	0.745	&	0.747	&	0.741	&	0.724	&	0.691	&	0.727	&	0.733	\\
SmBoP+GAZP	&	0.767	&	0.763	&	0.755	&	0.754	&	0.724	&	0.739	&	0.740	\\
SmBoP+GAZP+Preprocessing	&	0.754	&	0.748	&	0.743	&	0.742	&	0.691	&	0.733	&	0.728	\\
\hline
NatSmBoP.V.0	&	0.712	&	0.713	&	0.713	&	0.487	&	0.396	&	0.673	&	0.664	\\
NatSmBoP.V.1	&	0.704	&	0.703	&	0.704	&	0.495	&	0.408	&	0.667	&	0.660	\\
NatSmBoP.V.2	&	0.710	&	0.709	&	0.711	&	0.491	&	0.421	&	0.678	&	0.679	\\
NatSmBoP.V.0+Preprocessing	&	0.716	&	0.715	&	0.716	&	0.462	&	0.429	&	0.678	&	0.677	\\
NatSmBoP.V.1+Preprocessing	&	0.705	&	0.704	&	0.705	&	0.494	&	0.441	&	0.690	&	0.674	\\
NatSmBoP.V.2+Preprocessing	&	0.713	&	0.712	&	0.713	&	0.482	&	0.442	&	0.673	&	0.686	\\
\hline
RASAT	&	0.727	&	-	&	-	&	-	&	-	&	-	&	-	\\
NatRASAT	&	0.736	&	0.749	&	0.750	&	0.687	&	0.650	&	0.750	&	0.748	\\
\textbf{NatRASAT+Preprocessing}	&	0.757	&	0.771	&	\textbf{0.772}	&	0.718	&	0.686	&	0.761	&	0.757	\\
NatRASAT+GAZP	&	0.760	&	0.763	&	0.764	&	0.698	&	0.691	&	0.758	&	0.755	\\
NatRASAT+GAZP+Preprocessing	&	0.744	&	0.741	&	0.744	&	0.670	&	0.635	&	0.722	&	0.733	\\
 \hline
\end{tabular}}
\caption{Execution accuracies of models with different Rerankers trained using beams from SmBoP+GAZP}
\label{tab:ex-smbop-reranker}
\end{table*}

\begin{table*}[h]
\centering
\scalebox{0.83}{
\centering
\begin{tabular}{|l|c|cc|cc|cc|}
\hline
	&	\multirow{2}{*}{Baseline} &	\multicolumn{2}{c|}{SQL Verifier} &	\multicolumn{2}{c|}{SQL Selector} &	\multicolumn{2}{c|}{Binary SQL Selector} \\
 	&		& No Schema & \textbf{Schema} & Geometric & Random & Geometric & Random \\ \hline

SmBoP	&	0.750	&	0.751	&	0.751	&	0.565	&	0.428	&	0.664	&	0.669	\\
SmBoP+Preprocessing	&	0.745	&	0.744	&	0.744	&	0.579	&	0.440	&	0.686	&	0.700	\\
SmBoP+GAZP	&	0.767	&	0.766	&	0.766	&	0.627	&	0.475	&	0.690	&	0.700	\\
SmBoP+GAZP+Preprocessing	&	0.754	&	0.753	&	0.753	&	0.590	&	0.461	&	0.669	&	0.697	\\
\hline
NatSmBoP.V.0	&	0.712	&	0.702	&	0.691	&	0.466	&	0.261	&	0.563	&	0.544	\\
NatSmBoP.V.1	&	0.704	&	0.694	&	0.687	&	0.575	&	0.420	&	0.601	&	0.582	\\
NatSmBoP.V.2	&	0.710	&	0.695	&	0.688	&	0.590	&	0.487	&	0.639	&	0.605	\\
NatSmBoP.V.0+Preprocessing	&	0.716	&	0.699	&	0.694	&	0.574	&	0.422	&	0.627	&	0.618	\\
NatSmBoP.V.1+Preprocessing	&	0.705	&	0.699	&	0.692	&	0.601	&	0.509	&	0.653	&	0.607	\\
NatSmBoP.V.2+Preprocessing	&	0.713	&	0.703	&	0.697	&	0.600	&	0.449	&	0.613	&	0.608	\\
\hline
RASAT	&	0.727	&	-	&	-	&	-	&	-	&	-	&	-	\\
NatRASAT	&	0.736	&	0.757	&	0.752	&	0.751	&	0.736	&	0.753	&	0.742	\\
NatRASAT+Preprocessing	&	0.757	&	0.767	&	0.763	&	0.752	&	0.733	&	0.745	&	0.737	\\
\textbf{NatRASAT+GAZP}	&	0.760	&	0.766	&	\textbf{0.767}	&	0.757	&	0.738	&	0.754	&	0.750	\\
NatRASAT+GAZP+Preprocessing	&	0.744	&	0.760	&	0.754	&	0.745	&	0.742	&	0.750	&	0.738	\\
 \hline
\end{tabular}}
\caption{Execution accuracies of models with different Rerankers trained using beams from NatRASAT+GAZP}
\label{tab:ex-rasat-reranker}
\end{table*}

\begin{table*}[h]
\centering
\scalebox{0.83}{
\centering
\begin{tabular}{|l|c|cc|cc|cc|}
\hline
	& \multirow{2}{*}{Baseline} & \multicolumn{2}{c|}{SQL Verifier} & \multicolumn{2}{c|}{SQL Selector} &	\multicolumn{2}{c|}{Binary SQL Selector} \\
 	&		& No Schema & \textbf{Schema} & Geometric & Random & Geometric & Random \\ \hline

SmBoP	&	0.749	&	0.745	&	0.741	&	0.722	&	0.660	&	0.710	&	0.724	\\
SmBoP+Preprocessing	&	0.745	&	0.730	&	0.735	&	0.709	&	0.674	&	0.711	&	0.716	\\
SmBoP+GAZP	&	0.767	&	0.760	&	0.754	&	0.749	&	0.720	&	0.740	&	0.735	\\
SmBoP+GAZP+Preprocessing	&	0.754	&	0.757	&	0.752	&	0.749	&	0.696	&	0.736	&	0.737	\\
\hline
NatSmBoP.V.0	&	0.723	&	0.724	&	0.724	&	0.522	&	0.394	&	0.681	&	0.676	\\
NatSmBoP.V.1	&	0.723	&	0.722	&	0.723	&	0.530	&	0.458	&	0.698	&	0.683	\\
NatSmBoP.V.2	&	0.723	&	0.722	&	0.723	&	0.521	&	0.476	&	0.693	&	0.697	\\
NatSmBoP.V.0+Preprocessing	&	0.726	&	0.725	&	0.726	&	0.499	&	0.481	&	0.681	&	0.691	\\
NatSmBoP.V.1+Preprocessing	&	0.729	&	0.728	&	0.729	&	0.505	&	0.486	&	0.709	&	0.699	\\
NatSmBoP.V.2+Preprocessing	&	0.733	&	0.732	&	0.733	&	0.522	&	0.479	&	0.696	&	0.702	\\
\hline
RASAT	&	0.727	&	-	&	-	&	-	&	-	&	-	&	-	\\
NatRASAT	&	0.735	&	0.748	&	0.749	&	0.682	&	0.629	&	0.746	&	0.745	\\
\textbf{NatRASAT+Preprocessing}	&	0.762	&	0.764	&	\textbf{0.766}	&	0.731	&	0.703	&	0.756	&	0.752	\\
NatRASAT+GAZP	&	0.758	&	0.761	&	0.762	&	0.706	&	0.650	&	0.757	&	0.751	\\
NatRASAT+GAZP+Preprocessing	&	0.742	&	0.747	&	0.750	&	0.672	&	0.640	&	0.725	&	0.741	\\
 \hline
\end{tabular}}
\caption{Exact Match accuracy of models with different Rerankers trained using beams from SmBoP+GAZP}
\label{tab:em-smbop-reranker}
\end{table*}

\begin{table*}[h]
\centering
\scalebox{0.83}{
\centering
\begin{tabular}{|l|c|cc|cc|cc|}
\hline
	& \multirow{2}{*}{Baseline} &	\multicolumn{2}{c|}{SQL Verifier} &	\multicolumn{2}{c|}{SQL Selector} &	\multicolumn{2}{c|}{Binary SQL Selector} \\
 	&		& No Schema & \textbf{Schema} & Geometric & Random & Geometric & Random \\ \hline

SmBoP	&	0.749	&	0.746	&	0.746	&	0.538	&	0.394	&	0.647	&	0.662	\\
SmBoP+Preprocessing	&	0.745	&	0.732	&	0.732	&	0.563	&	0.419	&	0.662	&	0.693	\\
SmBoP+GAZP	&	0.767	&	0.766	&	0.766	&	0.600	&	0.446	&	0.672	&	0.696	\\
SmBoP+GAZP+Preprocessing	&	0.754	&	0.760	&	0.760	&	0.586	&	0.416	&	0.669	&	0.696	\\
\hline
NatSmBoP.V.0	&	0.723	&	0.715	&	0.700	&	0.458	&	0.238	&	0.580	&	0.558	\\
NatSmBoP.V.1	&	0.723	&	0.717	&	0.708	&	0.602	&	0.443	&	0.639	&	0.621	\\
NatSmBoP.V.2	&	0.723	&	0.716	&	0.702	&	0.620	&	0.525	&	0.666	&	0.631	\\
NatSmBoP.V.0+Preprocessing	&	0.726	&	0.716	&	0.708	&	0.587	&	0.439	&	0.651	&	0.634	\\
NatSmBoP.V.1+Preprocessing	&	0.729	&	0.718	&	0.709	&	0.625	&	0.535	&	0.686	&	0.632	\\
NatSmBoP.V.2+Preprocessing	&	0.733	&	0.727	&	0.716	&	0.625	&	0.485	&	0.644	&	0.632	\\
\hline
RASAT	&	0.727	&	-	&	-	&	-	&	-	&	-	&	-	\\
NatRASAT	&	0.735	&	0.761	&	0.750	&	0.759	&	0.735	&	0.760	&	0.747	\\
NatRASAT+Preprocessing	&	0.762	&	0.769	&	0.765	&	0.758	&	0.745	&	0.753	&	0.744	\\
\textbf{NatRASAT+GAZP}	&	0.758	&	0.769	&	\textbf{0.771}	&	0.765	&	0.753	&	0.762	&	0.758	\\
NatRASAT+GAZP+Preprocessing	&	0.742	&	0.766	&	0.759	&	0.747	&	0.747	&	0.756	&	0.743	\\
 \hline
\end{tabular}}
\caption{Exact Match accuracy of models with different Rerankers trained using beams from NatRASAT+GAZP}
\label{tab:em-rasat-reranker}
\end{table*}

\begin{table*}[htbp]
    \centering
    \begin{tabular}{l|ccccc|ccccc}	
\hline
\textbf{Metric} & \multicolumn{5}{c|}{Execution} & \multicolumn{5}{c}{Exact Match} \\
\textbf{Difficulty}	& ALL	&	easy	&	medium	&	hard	&	extra	&	ALL	&	easy	&	medium	&	hard	&	extra	\\ \hline
\textbf{System A}	& \textbf{0.772}	&	0.907	&	0.805	&	0.695	&	0.56	&	0.766	&	0.903	&	0.816	&	0.655	&	0.542	\\
\textbf{System B}	& 0.767	&	0.879	&	0.807	&	0.678	&	0.584	&	\textbf{0.771}	&	0.891	&	0.818	&	0.707	&	0.53	\\ \hline
    \end{tabular}
    \caption{Evaluation results of our two best models broken down by SQL difficulty (obtained on Spider dev set). System A: RASAT + NatSQL + Preprocessing + Schema-based SQL Verifier (trained on data from SmBoP+GAZP model); and System B: RASAT + NatSQL + GAZP + Schema-based SQL Verifier (trained on data from NatRASAT+GAZP model).}
    \label{t:best_models-category}
\end{table*}

\end{document}